\documentclass[11pt,a4paper]{article}
\usepackage[hyperref]{acl2020}
\usepackage{times}
\usepackage{latexsym}

\usepackage{microtype}

\aclfinalcopy 
\def\aclpaperid{1973} 

\usepackage{mathtools}
\usepackage{booktabs}
\usepackage{tikz}
\usepackage{pgfplots}
\usetikzlibrary{patterns}
\usepackage{multirow}
\usetikzlibrary{shapes}
\usetikzlibrary{decorations.pathreplacing,angles,quotes}

\newcommand{\egrid}{\mbox{EntityGrid}} %
\newcommand{\dd}{\mbox{DailyDialog}} %
\newcommand{\sw}{\mbox{SwitchBoard}}
\newcommand{\uolong}{Utterance Ordering (UO)}
\newcommand{\uoshort}{UO}
\newcommand{\uilong}{Utterance Insertion (UI)}
\newcommand{\uishort}{UI}
\newcommand{\urlong}{Utterance Replacement (UR)}
\newcommand{\urshort}{UR}
\newcommand{\huolong}{Even Utterance Ordering (EUO)}
\newcommand{\huoshort}{EUO}

\newcommand{\random}{\mbox{Random}}
\newcommand{\cosine}{\mbox{CoSim}}
\newcommand{\aseq}{\mbox{ASeq}}
\newcommand{\eagrid}{\mbox{EAGrid}}
\newcommand{\sdicoh}{\mbox{S-DiCoh}}
\newcommand{\mdicoh}{\mbox{M-DiCoh}}

\newcommand{\experiment}{\mbox{problem-domain}}
\newcommand{\annotation}{label}

\title{Dialogue Coherence Assessment \\ Without Explicit Dialogue Act Labels}

\author{
Mohsen Mesgar\And
Sebastian B{\"u}cker\vspace{0.2cm}\\
Ubiquitous Knowledge Processing Lab (UKP) \\
Technische Universit{\" a}t Darmstadt (TUDa)\\
\texttt{\{mesgar,buecker,gurevych\}@ukp.tu-darmstadt.de} \\\And
Iryna Gurevych \\
}

\date{}

\begin{document}
\maketitle

\begin{abstract}
Recent dialogue coherence models use the coherence features designed for monologue texts, e.g. nominal entities, to represent utterances and then explicitly augment them with dialogue-relevant features, e.g., dialogue act  \annotation s.
It indicates two drawbacks, (a) semantics of utterances is limited to entity mentions, and (b) the performance of coherence models strongly relies on the quality of the input dialogue act \annotation s.  
We address these issues by introducing a novel approach to dialogue coherence assessment. 
We use dialogue act prediction as an auxiliary task in a multi-task learning scenario to obtain informative utterance representations for coherence assessment.  
Our approach alleviates the need for explicit dialogue act \annotation s during evaluation.    
The results of our experiments show that our model substantially (more than $20$ accuracy points) outperforms its strong competitors on the DailyDialogue corpus, and performs on par with them on the \sw\ corpus for ranking dialogues concerning their coherence. 
We release our source code\footnote{\url{https://github.com/UKPLab/acl2020-dialogue-coherence-assessment}}.

\end{abstract}
\section{Introduction}
Considering rapid progresses in developing \mbox{open-domain} dialogue agents \cite{serban16,ghazvininejad18,dian18,zimingli19}, the need for models that compare these agents in various dialogue aspects becomes extremely important \cite{liuciawei16,dian18}. 
Most available methods for dialogue evaluation rely on word-overlap metrics, e.g. BLEU, and manually collected human feedback.  
The former does not strongly correlate with human judgments \cite{liuciawei16}, and the latter is time-consuming and subjective.  
A fundamental aspect of dialogue is coherence -- what discriminates a high-quality dialogue from a random sequence of dialogue utterances  \cite{halliday76,grosz86,byron98}. 
Dialogue coherence deals with semantic relations between utterances considering their dialogue acts \cite{perrault78,cervone18}. 

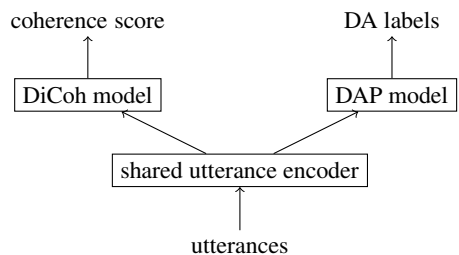
\begin{figure}[!t]
    \small
    \centering
    \begin{tikzpicture}

\node[] (utts) at (0,0){utterances};

\node[draw] (utt_module) at (0,1){shared utterance encoder};

\node[draw] (DAP_model) at (2,2){DAP model};

\node[] (dap_labels) at (2,3){DA labels};

\node[draw] (coh_model) at (-2,2){DiCoh model};

\node[] (coh_score) at (-2,3){coherence score};

\draw[->] (utts) -- (utt_module);
\draw[->] (utt_module) -- (DAP_model);
\draw[->] (utt_module) -- (coh_model);
\draw[->] (DAP_model) -- (dap_labels);
\draw[->] (coh_model) -- (coh_score);

\end{tikzpicture}
    \caption{A high-level view of our multi-task learning approach for dialogue coherence modeling.} 
    \label{fig:high_level_model}
\end{figure}

A Dialogue Act (henceforth \emph{DA}) gives a meaning to an utterance in a dialogue at the level of ``illocutionary force'', and therefore, constitutes the basic unit of communication \cite{searle69,raheja19}. 
A DA captures what a speaker's intention is of saying an utterance without regard to the actual content of the utterance. 
For example, a DA may indicate whether the intention of stating an utterance is to ask a question or to state a piece of information.

Recent approaches to dialogue coherence modeling use the coherence features designed for monologue texts, e.g. entity transitions \cite{barzilay05a}, and augment them with dialogue-relevant features, e.g., DA  \annotation s \cite{cervone18}.
These DA labels are provided by human annotators or DA prediction models. 
Such coherence models suffer from the following drawbacks: 
(a) they curb semantic representations of utterances to entities, which are sparse in dialogue because of short utterance lengths, and
(b) 
their performance relies on the quality of their input DA labels. 

We propose a novel approach to dialogue coherence assessment by  utilizing dialogue act prediction as an auxiliary task for training our coherence model in a multi-task learning (MTL) scenario (Figure~\ref{fig:high_level_model}). 
Our approach consists of three high-level components: \emph{an utterance encoder}, \emph{a dialogue coherence model (DiCoh)}, and \emph{a Dialogue Act Prediction (DAP) model}.  
The layers of the utterance encoder are shared between the DAP and the DiCoh model. 
This idea enables our DiCoh model to learn to focus on salient information presented in utterances considering their DAs and to alleviate the need for explicit DA labels during coherence assessment.  

We evaluate our MTL-based approach on the \dd\ \cite{liyanran17} and \sw\ \cite{jurafsky97} English dialogue corpora in several discriminating experiments, where our coherence model, DiCoh, is examined to  discriminate a dialogue from its perturbations (see Table~\ref{tab:motivation_example}). 
We utilize perturbation methods, like \emph{utterance ordering} and \emph{utterance insertion}, inherited from coherence evaluation approaches for monologue texts, and also introduce two dialogue-relevant perturbations, named \emph{utterance replacement} and \emph{even utterance ordering}. 
Our core contributions are: 
(1) proposing an MTL-based approach for dialogue coherence assessment using DAP as an auxiliary task, yielding more informative utterance representations for coherence assessment;
(2) alleviating the need for DA labels for dialogue coherence assessment during evaluations;
(3) an empirical evaluation on two benchmark dialogue corpora, showing that our model substantially outperforms the state-of-the-art coherence model on \dd, and performs on par with it on \sw.   

\section{Related Work}
\label{sec:rel}
Early approaches to dialogue coherence modeling are built upon available models for monologue, such as  the \egrid\ model \cite{barzilay05a,barzilay08}.
\egrid\ and its extensions \cite{burstein10,guinaudeau13,mesgar14,nguyen17,farag19} rely on entity transitions, as proxies of semantic connectivity, between utterances.  
These approaches are agnostic to discourse properties of dialogues \cite{purandare08,gandhe08,cervone18}. 
\begin{table}[!t]
    \centering
    \small
    \begin{tabular}{@{}l@{}l@{}l@{}}
    \toprule
        & Utterance & DA label\\
    \midrule
        \textbf{coherent} \\
         $utt_1$: &  \textit{This is my uncle, Charles.} & inform \\
         $\mathbf{utt_2}$ &  \textbf{\textit{He looks strong. What does he do? }} & question\\
         $utt_3$: & \textit{He's a captain.} & inform \\
         $\mathbf{utt_4}$: &  \textbf{\textit{He must be very brave.}} & inform\\
         $utt_5$: & \textit{Exactly!} & inform\\
         \midrule
         \textbf{incoherent} \\
         $utt_1$:: &  \textit{This is my uncle, Charles.} & inform\\
          $\mathbf{utt_4}$: & \textbf{\textit{He must be very brave.}} & inform\\
         $utt_3$: & \textit{He's a captain.} & inform \\
         $\mathbf{utt_2}$: & \textbf{\textit{He looks strong. What does he do? }} & question\\
          $utt_5$: & \textit{Exactly!} & inform\\
    \bottomrule
    \end{tabular}
    \caption{An example dialogue from \dd\ (top) and its perturbation (bottom), which is generated by permuting the utterances said by one of the speakers (shown in boldface), and is less coherent. 
    The right column shows the DA labels associated with utterances.}
    \label{tab:motivation_example}
\end{table}

Inspired by \egrid, \newcite{gandhe16} define transition patterns among DA \annotation s associated with utterances to measure coherence. 
\newcite{cervone18} combine the above ideas by augmenting entity grids with utterance DA labels. 
This model restricts utterance vectors only to  entity mentions, and needs gold DA labels as its inputs for training as well as evaluation. 
However, obtaining DA labels from human annotators is expensive and using DAP models makes the performance of coherence model dependent on the performance of DAP models. 

Recent approaches to dialogue coherence modeling benefit from distributional representations of utterances. 
\newcite{zhanghainan18} quantify the coherence of dialogue using the semantic similarity between each utterance and its preceding utterances. 
This similarity is estimated, for example, by the cosine similarity between an utterance vector and a context vector where those vectors are the average of their pre-trained word embeddings. 
\newcite{vakulenko18} measure dialogue coherence based on the consistency of new concepts introduced in a dialogue with background knowledge. 
Similarly, \newcite{dziri19} utilize a natural language inference model to assess the content consistency among utterances as an indicator for dialogue coherence. 
However, these approaches lack dialogue-relevant information to measure coherence. 
Our MTL-based approach solves these issues: 
(i) it benefits from DAs and semantics of utterances to measure dialogue coherence by optimizing utterance vectors for both DAP and coherence assessment, and 
(ii) it uses DA labels to define an auxiliary task for training the DiCoh model using MTL, instead of utilizing them in a pipeline. Therefore, it efficiently mitigates the need for explicit DA labels as inputs during coherence assessment. 

\begin{figure*}[!t]
    \small
    \centering
    \resizebox{\textwidth}{5.3cm}{%
    \begin{tabular}{cc}
\multicolumn{2}{c}{
\begin{tikzpicture}
\node[] (n0) at (0,0){};
\node[draw, ellipse,dashed] (hing) at (0,1){preference loss};
\node[] (l-coh) at (0,2){$\mathcal{L}^p_{coh}$};

\draw[-,dashed] (hing) -- (l-coh);
\draw[-,dashed] (-3,0) -- (hing);
\draw[-,dashed] (+3,0) -- (hing);
\end{tikzpicture}
}
\\ 
\begin{tikzpicture}
\node[] (n0) at (0,0){};

\node[] (utt-0) at (0,0){$utt_1=[w_1,...,w_n],$};
\node[draw] (emb-layer-0) at (0,1){Emb};
\node[] (emb-0) at (0,2){$[e_1,...,e_n]$};
\node[draw] (bi-lstm-0) at (0,3){BiLSTM};
\node[] (hidden-0) at (0,4){$[h^u_1,...,h^u_n]$};
\node[draw] (atten-0) at (0,5){Atten};
\node[] (u-0) at (0,6){$u_1$};

\draw[-] (utt-0) -- (emb-layer-0);
\draw[-] (emb-layer-0) -- (emb-0);
\draw[-] (emb-0) -- (bi-lstm-0);
\draw[-] (bi-lstm-0) -- (hidden-0);
\draw[-] (hidden-0) -- (atten-0);
\draw[-] (atten-0) -- (u-0);

\node[] (middle-utts) at (2,0) {$ ... $};
\node[] (middle-utts) at (2,3) {$ ... $};
\node[] (middle-utts) at (2,6) {$ ... $};

\node[] (utt-m) at (4,0){$utt_m=[w_1,...,w_n],$};
\node[draw] (emb-layer-m) at (4,1){Emb};
\node[] (emb-m) at (4,2){$[e_1,...,e_n]$};
\node[draw] (bi-lstm-m) at (4,3){BiLSTM};
\node[] (hidden-m) at (4,4){$[h^u_1,...,h^u_n]$};
\node[draw] (atten-m) at (4,5){Atten};
\node[] (u-m) at (4,6){$u_m$};

\draw[-] (utt-m) -- (emb-layer-m);
\draw[-] (emb-layer-m) -- (emb-m);
\draw[-] (emb-m) -- (bi-lstm-m);
\draw[-] (bi-lstm-m) -- (hidden-m);
\draw[-] (hidden-m) -- (atten-m);
\draw[-] (atten-m) -- (u-m);

\node[draw, minimum width=5cm] (di-bi-lstm) at (2,7) {BiLSTM};
\node[] (di-hidden-d) at (2,8) {$[h^d_1,...,h^d_m]$};
\node[draw] (atten-d) at (2,9) {Atten};
\node[] (d) at (2,10) {$d$};

\draw[-] (u-0) -- (0,6.7);
\draw[-] (u-m) -- (4,6.7);
\draw[-] (di-bi-lstm) -- (di-hidden-d);
\draw[-] (di-hidden-d) -- (atten-d); 
\draw[-] (atten-d) -- (d);

\node[draw] (proj) at (2,11) {Linear};
\node[] (s-d) at (2,12) {$s_{dial_i}$};
\draw[-] (d) -- (proj);
\draw[-] (proj) -- (s-d);

\node[draw, minimum width=4cm] (da-classifier) at (-3,7) {softmax}; 
\draw[-] (0,6.6) -- (-4.5,6.6);
\draw[-] (-4.5,6.6) -- (-4.5,6.7);
\node[] (middle-utts) at (-3,6.5) {$...$}; 
\draw[-] (4,6.4) -- (-2.0,6.4);
\draw[-] (-2.0,6.4) -- (-2.0,6.7);

\node[] (a-0) at (-4.5,8) {$a_1$}; 
\node[] (middle-acts) at (-3,8) {$...$}; 
\node[] (a-m) at (-2.0,8) {$a_m$};
\draw[-] (-4.5,7.2) -- (a-0);
\draw[-] (-2.0,7.2) -- (a-m);
\draw[decoration={brace,mirror,raise=5pt},decorate]
  (-1.5,-0.5) -- node [below=6pt] {$dial_i $} (5.5,-0.5);

\node[draw, ellipse,dashed] (avg-xent) at (-3,9) {avg. cross-entropy};
\draw[-,dashed] (a-0) -- (-4.5,8.5);
\draw[-,dashed] (a-m) -- (-2.0,8.5);

\node[] (l-da-i) at (-3,10) {$\mathcal{L}^{dial_i}_{da}$};
\draw[-,dashed] (avg-xent) -- (l-da-i);

\end{tikzpicture}

& 

\begin{tikzpicture}
\node[] (n0) at (0,0){};

\node[] (utt-0) at (0,0){$utt_1=[w_1,...,w_n],$};
\node[draw] (emb-layer-0) at (0,1){Emb};
\node[] (emb-0) at (0,2){$[e_1,...,e_n]$};
\node[draw] (bi-lstm-0) at (0,3){BiLSTM};
\node[] (hidden-0) at (0,4){$[h^u_1,...,h^u_n]$};
\node[draw] (atten-0) at (0,5){Atten};
\node[] (u-0) at (0,6){$u_1$};

\draw[-] (utt-0) -- (emb-layer-0);
\draw[-] (emb-layer-0) -- (emb-0);
\draw[-] (emb-0) -- (bi-lstm-0);
\draw[-] (bi-lstm-0) -- (hidden-0);
\draw[-] (hidden-0) -- (atten-0);
\draw[-] (atten-0) -- (u-0);

\node[] (middle-utts) at (2,0) {$ ... $};
\node[] (middle-utts) at (2,3) {$ ... $};
\node[] (middle-utts) at (2,6) {$ ... $};

\node[] (utt-m) at (4,0){$utt_m=[w_0,...,w_n],$};
\node[draw] (emb-layer-m) at (4,1){Emb};
\node[] (emb-m) at (4,2){$[e_1,...,e_n]$};
\node[draw] (bi-lstm-m) at (4,3){BiLSTM};
\node[] (hidden-m) at (4,4){$[h^u_1,...,h^u_n]$};
\node[draw] (atten-m) at (4,5){Atten};
\node[] (u-m) at (4,6){$u_m$};

\draw[-] (utt-m) -- (emb-layer-m);
\draw[-] (emb-layer-m) -- (emb-m);
\draw[-] (emb-m) -- (bi-lstm-m);
\draw[-] (bi-lstm-m) -- (hidden-m);
\draw[-] (hidden-m) -- (atten-m);
\draw[-] (atten-m) -- (u-m);

\draw[decoration={brace,mirror,raise=5pt},decorate]
  (-1.5,-0.5) -- node [below=6pt] {$dial_j$} (5.5,-0.5);

\node[draw, minimum width=5cm] (di-bi-lstm) at (2,7) {BiLSTM};
\node[] (di-hidden-d) at (2,8) {$[h^d_1,...,h^d_m]$};
\node[draw] (atten-d) at (2,9) {Atten};
\node[] (d) at (2,10) {$d$};

\draw[-] (u-0) -- (0,6.7);
\draw[-] (u-m) -- (4,6.7);
\draw[-] (di-bi-lstm) -- (di-hidden-d);
\draw[-] (di-hidden-d) -- (atten-d); 
\draw[-] (atten-d) -- (d);

\node[draw] (proj) at (2,11) {Linear};
\node[] (s-d) at (2,12) {$s_{dial_j}$};
\draw[-] (d) -- (proj);
\draw[-] (proj) -- (s-d);

\node[draw, minimum width=4cm] (da-classifier) at (7,7) {softmax}; 
\draw[-] (0,6.6) -- (5.5,6.6);
\draw[-] (5.5,6.6) -- (5.5,6.7);
\node[] (middle-utts) at (7,6.6) {$...$}; 
\draw[-] (4,6.4) -- (8.5,6.4);
\draw[-] (8.5,6.4) -- (8.5,6.7);

\node[] (a-0) at (5.5,8) {$a_1$}; 
\node[] (middle-acts) at (7,8) {$...$}; 
\node[] (a-m) at (8.5,8) {$a_m$};
\draw[-] (5.5,7.2) -- (a-0);
\draw[-] (8.5,7.2) -- (a-m);

\node[draw, ellipse,dashed] (avg-xent) at (7,9) {avg. cross-entropy};
\draw[-,dashed] (a-0) -- (5.5,8.5);
\draw[-,dashed] (a-m) -- (8.5,8.5);

\node[] (l-da-j) at (7,10) {$\mathcal{L}^{dial_j}_{da}$};
\draw[-,dashed] (avg-xent) -- (l-da-j);

\end{tikzpicture}

\end{tabular}
    }
    \caption{A low-level illustration of our MTL-based approach to dialogue coherence assessment. The input is dialogue pair $p=(dial_i, dial_j)$. Dashed items represent losses. Models' parameters are shared among dialogues.}
    \label{fig:model}
\end{figure*}
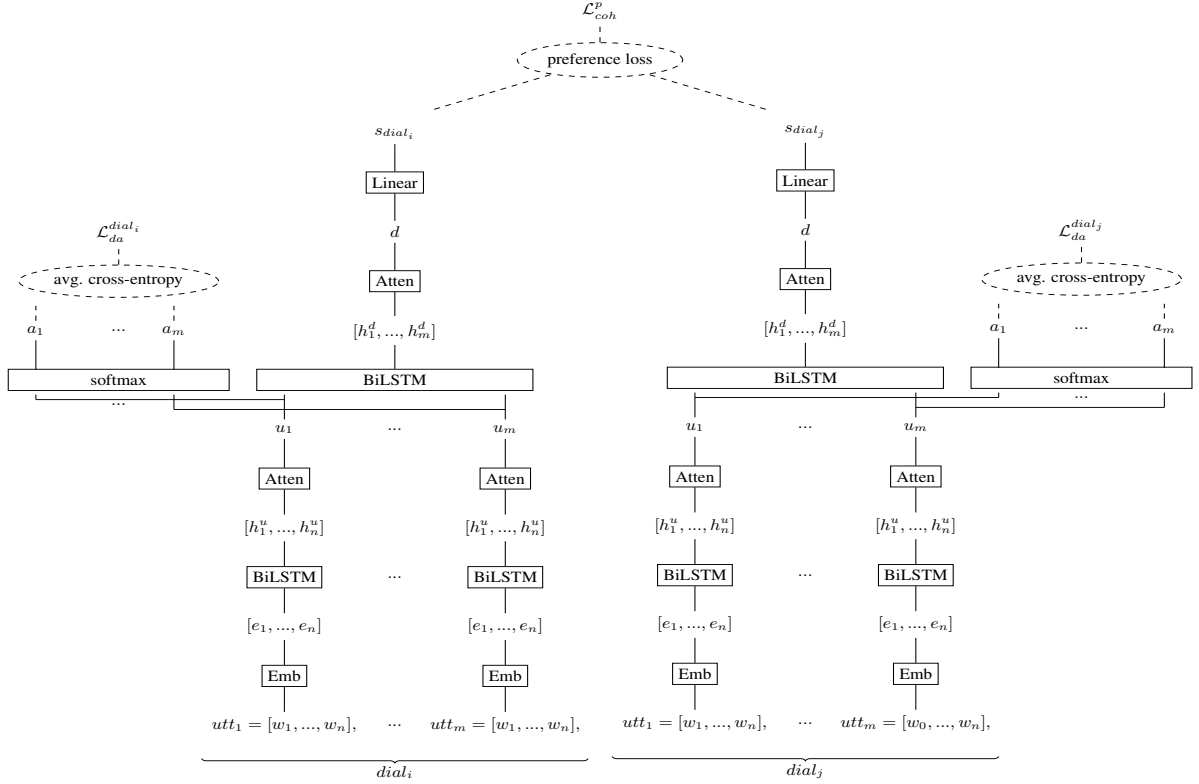 
%
\section{Method}
We represent a dialogue between two speakers as a sequence of utterances, \mbox{$\textit{dial} = [\textit{utt}_1,...,\textit{utt}_m]$}. 
We address the problem of designing a coherence model, DiCoh, which assigns a coherence score to $\textit{dial}$, \mbox{$s_{\textit{dial}} = $\textit{DiCoh}$(\textit{dial})$}. 
Given a pair of dialogues \mbox{$\phi=(\textit{dial}_i,\textit{dial}_j)$}, our $\textit{DiCoh}$ model ideally assigns $s_{\textit{dial}_i} > s_{\textit{dial}_j}$ if and only if dialogue $\textit{dial}_i$ is preferred over dialogue $\textit{dial}_j$ according to their perceived coherence. 
Instead of using gold DA labels as inputs to DiCoh, we use them to define an auxiliary task and model, DAP, to enrich utterance vectors for DiCoh in an MTL scenario. 
Figure~\ref{fig:model} shows a low-level illustration of our MTL-based approach.    

\paragraph{Utterance encoder} 
We use a word embedding layer, $\textit{Emb}$, to transform the words in utterance $\textit{utt} = [w_1,..., w_n]$ to a sequence of embedding vectors $E = [e_1,...,e_n]$, where $n$ is the number of words in $\textit{utt}$.  
The embedding layer can be initialized by any pre-trained embeddings to capture lexical relations.    
We use a Bidirectional recurrent neural network with Long Short-Term Memory cells, $\textit{BiLSTM}$, to map embeddings $E$ to encode words in their utterance-level context: 
\begin{equation}
    \begin{split}
        E &=   \textit{Emb}(\textit{utt}), \\
        H_u &= \textit{BiLSTM}(E),
    \end{split}
    \label{eq:lstm}
\end{equation}
where $H_u$ shows the hidden state vectors $[h^u_1,..., h^u_n]$ returned by $\textit{BiLSTM}$. 
At word $t$, $h^u_t$ is the concatenation of hidden states of the forward $\overrightarrow{h^u_t}$ and the backward LSTMs $\overleftarrow{h^u_t}$:  
\begin{equation}
    \begin{split}
        h^u_t &= [\overrightarrow{h^u_t};\overleftarrow{h^u_t}].
    \end{split}
    \label{eq:concat}
\end{equation}
We apply a self-attention mechanism, $\textit{Atten}$, to the hidden state vectors in $H_u$ to obtain the vector representation, $u$, of utterance $utt$:
\begin{equation}
     u = \textit{Atten}(H_u).
    \label{eq:selfatten}
\end{equation}
Generally, the attention layer, $\textit{Atten}$, for an input vector $x$ is defined as follows:
\begin{equation}
            \begin{split}
            \beta_t & = x_t* W,  \\
            \alpha_t & = \frac{\exp{(\beta_t)}}{\sum_t \exp{(\beta_t)}}, \\
            o & = \sum_t \alpha_t * x_t,
        \end{split}
        \label{eq:attn}
\end{equation}
where $W$ is the parameter of this layer, and $o$ is its weighted output vector. 
Attention enables the utterance representation layer to encode an utterance by the weighted sum of its word embeddings.  
It is worth noting that the parameters of the utterance encoder are shared for representing all utterances in a dialogue. 
\paragraph{DiCoh model}
For an input dialogue \mbox{$\textit{dial}=[\textit{utt}_1,...,\textit{utt}_m]$}, the output of the utterance representation encoder is a sequence of vectors, i.e., $[u_1,..., u_m]$. 
Our coherence assessment model (DiCoh) combines these vectors by a $\textit{BiLSTM}$ to obtain dialogue-level contextualized representations of utterances. 
Then, a self-attention (Equation~\ref{eq:attn}) with new parameters computes the weighted average of the contextualized utterance vectors to encode the dialogue:
\begin{equation}
    \begin{split}
         [h^d_1,..., h^d_m] &= \textit{BiLSTM}([u_1,...,u_m]), \\
          d &= \textit{Atten}([h^d_1,..., h^d_m]).
    \end{split}
    \label{eq:dialogue}
\end{equation}
A linear feed-forward layer, $FF$, maps the dialogue vector, $d$, to a dialogue coherence score, $s_{dial}$:
\begin{equation}
    \begin{split}
          s_{\textit{dial}} &= \textit{FF}(d).
    \end{split}
    \label{eq:dialogue}
\end{equation}
\paragraph{DAP model} 
%
Our DAP model, which is used to solve the auxiliary DAP task, is a $softmax$ layer which maps an utterance vector, $u$, to a probability distribution $p_a$ over DA \annotation s $A$: 
\begin{equation}
    \begin{split}
       p_a(u) &= \textit{softmax}(W_{|u|\times |A|}*u + b),\\
    \end{split}
    \label{eq:da-classification}
\end{equation}
where $W_{|u|\times |A|}$ shows the weights of the $\textit{softmax}$ layer, $|u|$ is the size of the utterance vector, $|A|$ is the number of DA labels, and $b$ is the bias. 

\subsection{Multi-Task Learning}
As illustrated in Figure~\ref{fig:high_level_model}, our main idea is to benefit from the DAP task for improving the performance of the dialogue coherence model by using them in a multi-task learning scenario. 
%
We also assume that each utterance $utt_k$ is associated with DA label, $a_k$, during training but not during evaluation.  

We define a loss function for each task, and then use their weighted average as the total loss.  
The DAP loss function for dialogue $dial$ is the average \mbox{cross-entropy}: 
\begin{equation}
    \mathcal{L}^{\textit{dial}}_{da} = -\frac{1}{m}{\sum_{k\in(1,...,m)} }a_k*\log(p_a(u_k)),
\end{equation}
where $m$ is the number of utterances in dialogue, and $a_k$ is the one-hot vector representation of the gold DA \annotation\ associated with the $k^{th}$ utterance.
$\textit{log}(p_a)$ is the natural log of probabilities over DA labels, which is obtained in Equation~\ref{eq:da-classification}. 

Inspired by preference learning approaches (e.g.\ the proposed method by \newcite{gaoyang19} for text summarization) we define the loss function for coherence assessment through pairwise comparisons among dialogues. 
Given dialogue pair $\phi = (\textit{dial}_i,\textit{dial}_j)$ and its preference coherence label, 
\begin{equation}
l^c =
\left\{
	\begin{array}{ll}
	0  & \mbox{if } \textit{dial}_i \mbox{ is preferred over }\textit{dial}_j, \\
	1 & \mbox{otherwise,}
	\end{array}
\right.
\end{equation}
the coherence loss is: 
\begin{equation}
    \mathcal{L}^\phi_{\textit{coh}}= \max \lbrace 0, 1 - s_{\phi\left[l^c \right]} + s_{\phi\left[1-l^c\right]} \rbrace,
    \label{eq:coh-loss}
\end{equation}
where $ \left[.\right]$ is the indexing function. More formally, $s_{\phi \left[ l^c \right]}$ and $s_{\phi \left[ 1-l^c \right]}$ are the coherence scores of the coherent and incoherent dialogue in pair $\phi = (dial_i,dial_j)$, respectively.
Finally, the total loss value is the weighted combination \cite{kendall18} of the above losses: 
\begin{equation}
    \mathcal{L} = \frac{\mathcal{L}^\phi_{\textit{coh}}}{\gamma_1^2} + \frac{(\mathcal{L}^{\textit{dial}_i}_{da}+ \mathcal{L}^{\textit{dial}_j}_{da})}{\gamma_2^2} + \log(\gamma_1) +  \log(\gamma_2),
    \label{eq:total_loss}
\end{equation}
where $\mathcal{L}^{\textit{dial}_i}_{da}$ and $\mathcal{L}^{\textit{dial}_j}_{da}$ are the losses of DAP for dialogues in pair $\phi=(\textit{dial}_i, \textit{dial}_j)$, $\gamma_1$ and $\gamma_2$ are trainable parameters to balance the impact of losses.  
We compute the gradient of $\mathcal{L}$ to update the parameters of both DiCoh and DAP models.

\section{Experiments}

\subsection{Dialogue Corpora}
We compare our approach with several previous dialogue coherence models on \dd\ \cite{liyanran17} and \sw\ \cite{jurafsky97} as two benchmark English dialogue corpora.
%
Table~\ref{tab:datasets} shows some statistics of these corpora. 
\begin{table}[!t]
    \centering
    \small
    \begin{tabular}{@{}lll@{}}
     \toprule
         & \dd & \sw \\
    \midrule
        \# dialogues &  $13{,}118$ & $1{,}155$ \\
        \# DA labels & $4$ &  $42$\\
        avg. \# utter. per dialogue & $7.9$  & $191.9$  \\
        avg. \# words per utter. & $14.6$ &$9.26$  \\
    \bottomrule
    \end{tabular}
    \caption{The statistics of the \dd\ and \sw\ corpora.}
    \label{tab:datasets}
\end{table}

\dd\ contains human-written dialogues about daily topics (e.g.\ ordinary life, relationships, work, etc) collected by crowd-sourcing. 
Crowd-workers also annotated utterances with generic DA \annotation s from the set \{Inform, Question, Directive, Commissive\}.  
Dialogues in this corpus contain a few utterances ($\approx 8$) making them more on topic and less dispersed. 
However, utterances are long in terms of the number of words ($\approx 15$).     

\sw\ contains informal English dialogues collected from phone conversations between two mutually unknown human participants.  
The participants were given only one of $70$ possible topics as initial topic to start a conversation but they were free to diverge from that topic during the conversation. 
So, there is no concrete topic associated with each dialogue in this dataset as it is the case for dialogues in \dd . 

DA \annotation s in \sw\ are about $10$ times more fine-grained than those in \dd . 
For example, a question utterance in \sw\ may have a fine-grained DA label such as Yes-No-Question, Wh-Question, Rhetorical-Questions, etc.  
The distribution of these acts is however highly unbalanced in \sw : the most frequent act label makes up for 36\% of the utterances in the corpus, the three most frequent acts together make up for 68\% of the utterances, while most of the remaining act labels just make up for 1\% or less of all the utterances. 

On average, dialogues in \sw\ contain more utterances than those in \dd\ ($192$ vs $8$) but utterances in \sw\ are shorter than those in \dd\  ($9$ vs $15$).  
This means that dialogues in \sw\ are more likely to span different topics than the ones in \dd.
The utterances in \dd\ are explicitly cleaned of any noise, like ``uh-oh'', or interruptions by the other speaker, as it is commonly the case for dialogues in \sw . 
While each dialogue turn of dialogues in \dd\ contains only one utterance, dialogue turns in \sw\ may consist of several utterances.
That is why we consider each dialogue as a sequence of dialogue utterances. 

\subsection{Problem-domains} 
The goal of our experiments is to assess if a coherence model assigns coherence scores to  dialogues so that a more coherent dialogue obtains a higher score than a less coherent one. 
Since dialogues in the examined corpora, i.e. \dd\ and \sw\ , are not associated with any coherence assessment score, we synthetically define four perturbation methods to destroy the coherence of dialogues in these corpora, and create a set of dialogue pairs for training and testing coherence models.  

We borrow Utterance Ordering (UO) and Utterance Insertion (UI) from previous studies on coherence assessment \cite{barzilay05a,cervone18} and also  introduce Utterance Replacement (UR), and Even Utterance Ordering (EUO) as more challenging and dialogue-relevant perturbation methods.  
Since each experiment follows a specific perturbation method, henceforth, we refer to these perturbations as problem-domains: 
    \paragraph{\uolong} 
    We randomly permute the order of utterances in dialogue. 
    The original dialogue is preferred over the perturbed one.  
    \paragraph{\uilong} 
     We remove each utterance of a dialogue and then re-insert it in any possible utterance position in the dialogue. 
    We assume that the original place of the utterance is the best place for the insertion. 
    Therefore, a coherence model ideally discriminates the original dialogue from the perturbed ones, which are obtained by re-inserting the removed utterance in any utterance position except its original one. 
    This \experiment\ is more difficult to solve than \uoshort\ as the distinction between dialogues is in the position of only one utterance. 
    \paragraph{\urlong}
    We randomly replace one of the utterances in a dialogue with another utterance that is also  randomly selected from another dialogue. 
    The original dialogue is preferred over the dialogue generated by \urshort. 
    Unlike the other problem-domains, which perturb the structure of a dialogue, this problem-domain perturbs the coherence of a dialogue at its semantic level.   
    %
    \paragraph{\huolong}
    This \experiment\ is similar to \uoshort\ but here we re-arrange the order of utterances that are said by one speaker and keep the order of the other utterances, which are said by the other speaker, fixed. 
    Therefore, \huoshort\ is more challenging and  dialogue-relevant than \uoshort .  
    This problem-domain assesses to what extent coherence models capture the coherence among utterances that are said by one of the speakers in a dialogue.  

\subsection{Problem-domain Datasets}
To create dialogue pairs for each problem-domain, we use the splits provided by the \dd\ corpus; and for \sw\ we take $80\%$ of dialogues for the training, $10\%$ for the validation and $10\%$ for the test sets.  
Following \newcite{cervone18}, for any dialogue in each set we create $20$ perturbations where each of which makes two pairs with the original dialogue. 
Given dialogue $dial_i$ and its perturbation $dial_j$, we define two dialogue pairs: $(dial_i,dial_j)$ with preference coherence label \mbox{$l^c=0$} and $(dial_j,dial_i)$ with label $l^c=1$. 

\subsection{In problem-domain Evaluation}
In this evaluation, we train, fine-tune, and evaluate our models on the training, validation, and test sets of each \experiment. 
Note that these sets are constructed by the same perturbation method.    
\begin{table*}[!ht] 
    \centering
    \small
    \resizebox{\textwidth}{!}{
    \begin{tabular}{lcccc|cccc}
    \toprule
        & \multicolumn{4}{c}{\dd} & \multicolumn{4}{c}{\sw}\\
         Model & \uoshort & \uishort & \urshort & \huoshort & \uoshort & \uishort & \urshort & \huoshort \\
         \midrule
         \random  & $50.10$ & $49.97$ & $49.97$ & $49.92$ & $49.98$ & $50.02$ & $49.99$ & $50.13$ \\
         \cosine  & $57.20$ & $50.88$ & $65.18$ & $66.86$ & $82.84$ & $55.63$ & $50.87$ & $74.48$ \\
         \aseq    & $68.21$ & $57.41$ & $61.89$ & $62.73$ & $\mathbf{99.70}$ & $73.94$ & $63.48$ & $99.20$ \\ 
         \eagrid  & $71.72$ & $60.93$ &	$68.49$ & $67.18$ & $99.65$ & $73.70$ & $\mathbf{75.61}$ & $\mathbf{99.83}$ \\
         \midrule
         \sdicoh  & $94.23\pm .74$ & $83.33 \pm .81$ &	$81.89 \pm .26$ & $86.38 \pm .29$ & $95.51 \pm .61$ & $80.60 \pm 1.12$ & $53.61 \pm .35$ & $88.83 \pm .35$ \\
         \mdicoh  & $\mathbf{95.92 \pm .12}$ & $\mathbf{88.20 \pm .36}$ & $\mathbf{83.02 \pm .50}$ & $\mathbf{88.55 \pm .39}$ & $99.41 \pm .11$ & $\mathbf{85.04 \pm 1.14}$ & $58.67 \pm 1.79$     & $97.08 \pm .20$ \\ 
        \bottomrule
    \end{tabular}
    }
    \caption{The accuracy (\%) of the examined models on the test set of each experiment defined on \dd\ and \sw. }
    \label{tab:dd_res}
\end{table*}

\paragraph{Compared coherence models}
We compare the following coherence models in this evaluation: 
(1) \textbf{\random:}  This baseline model randomly ranks dialogues in an input dialogue-pair.   
(2) \textbf{\cosine} \cite{zhanghainan18,xuxinnuo18}\textbf{:} 
     This model represents utterances by averaging the pre-trained embeddings of their words. 
    Then, the average of the cosine similarities between vectors of adjacent utterances is taken as the coherence score. 
    In this model, utterance vectors are made using content words by eliminating all stop words. 
    (3) \textbf{\aseq} \cite{gandhe16}\textbf{:} 
    This model relies only DAs transitions and is agnostic to  semantic relationships (such as entity transitions) between utterances. 
    Coherence features in this model are the probabilities of n-grams across the sequence of DAs associated with the utterances in dialogue. 
    These features are supplied to a SVM to rank dialogues. 
    (4) \textbf{\eagrid} \cite{cervone18}\textbf{:}
    This is the best performing model presented by \newcite{cervone18} that benefits from both entity and DA transitions between utterances. 
    It represents semantic relationships across utterances via a grid, whose rows are associated with utterances and all columns represent entities but one that represents DAs. 
    Entities are a set of mentions that are extracted by a co-reference system. 
    Entries at the intersections between entity columns and an utterance row represent the grammatical role of an entity in an utterance.
    The intersection of the DA column and an utterance shows the DA label of the utterance.
    \newcite{cervone18} use grammatical role transitions of entities as well as DA label transitions across utterances as indicative patterns for coherence. 
    The frequencies of these patterns are taken as coherence features, which are supplied to Support Vector Machines (SVMs) to discriminate dialogues with respect to their coherence. 
    (5) \textbf{\sdicoh:} 
    This is our coherence model, DiCoh, trained by only the supervision signal for coherence ranking, with  
    the total loss $\mathcal{L}=\mathcal{L}^\phi_{coh}$ (see Equation~\ref{eq:total_loss}). 
    This model does not benefit from DA information to enrich utterance vectors. 
    (6) \textbf{\mdicoh:}
    This is our full model trained by the proposed MTL using the supervision signals for both coherence ranking and DAP.  
    The main advantage of this model is that it learns to focus on salient information of utterances for coherence assessment based on the given DAs for utterances. 

We follow former coherence papers \cite{barzilay08,guinaudeau13,mesgar18,cervone18} and use \emph{accuracy} as the evaluation metric. 
In our experiments, this metric equals the frequency of correctly discriminated dialogue pairs in the test set of a problem-domain.  
\begin{equation}
    acc = \frac{\text{\# of correctly discriminated dialogue pairs}}{\text{\# of dialogue pairs}}.
\end{equation}
To reduce the risk of randomness in our experiments, we run each experiment five times with varying random seeds and report their average \cite{reimers18}. 
\paragraph{Settings}
Each batch consists of $128$ and $16$ dialogue-pairs for the \dd\ and \sw\ corpora, respectively. 
Utterances are zero-padded and masked. 
We use pretrained GloVe embeddings \cite{pennington14} of size $300$ wherever word embeddings are required (i.e., in \cosine, \sdicoh, and \mdicoh). 
For the \cosine\ model, we use the SMART English stop word list \cite{salton71} to eliminate all stop words. 
For the \aseq\ model, we use bi-grams of DA \annotation s to define the coherence features \cite{cervone18}.   
All parameters of the \eagrid\ model have the same value as the best performing model proposed by \newcite{cervone18}.

In DiCoh, the size of the hidden states in LSTMs of the utterance module is $128$ and of the dialogue module is $256$.  
The parameters of this model are optimized using the Adam optimizer where its parameters have default values except the learning rate which is initiated with $0.0005$. 
A dropout layer with $p=0.1$ is applied to the utterance vectors. 
We train the model for $20$ epochs on \dd\ and $10$ epochs on \sw\ and evaluate it at the end of each epoch on the validation set.  
The best performing model on the validation set is used for the final evaluation on the test set.
Parameters $\gamma_1$ and $\gamma_2$ (see Equation~\ref{eq:total_loss}) are initiated with $2.0$ and are updated during training.  
To have fair comparisons, we train and evaluate all compared models on identical training, validation, and test sets. 
\paragraph{Results} 
Table~\ref{tab:dd_res} shows the accuracy of the baseline models (top) and our model (bottom) on \dd\ and \sw.     

We investigate how well our DiCoh model performs in comparison with its baseline peers that do not take DAs into account, i.e., \random\ and \cosine.  
We observe that \mbox{\sdicoh} strongly outperforms these models for all the examined \experiment s on both \dd\ and \sw, confirming the validity of our DiCoh model for capturing the semantics of utterances. 

In a more challenging comparison, we compare \sdicoh\ with \aseq\ and \eagrid\ as the baseline models that use DA information. 
Our \sdicoh\ even surpasses these models for all \experiment s on \dd.  
However, on \sw, \sdicoh\ achieves lower accuracy than these models for all \experiment s except \uishort.
This observation shows that when dialogue utterances are short (like those in \sw\ in comparison with those in \dd), DAs are more crucial for coherence assessment. 
It is worth noting that unlike \eagrid\ and \aseq , \sdicoh\ is completely agnostic to DA information. 

When we employ DAP as an auxiliary task to train the DiCoh model in our MTL setup, we observe that 
%
\mdicoh\ substantially outperforms the \random,  \cosine, and \sdicoh\ models (which do not use DAs) for all \experiment s on both \dd\ and \sw. 
It concludes that our proposed MTL approach effectively leverages the DAP task to learn informative utterance vectors for dialogue coherence assessment. 

Compared with the \aseq\ and \eagrid\ models, which explicitly use gold DA labels during evaluations,  our \mdicoh\ achieves the highest accuracy for all \experiment s on \dd, showing that our approach for involving DAs yields more informative utterance representations for coherence assessments. 
However, on \sw, \mdicoh\ increases the accuracy of \sdicoh\ up to those of \eagrid\ for \uoshort\ and \huoshort.    
Surprisingly, it achieves lower accuracy than what \eagrid\ achieves for \urshort.  

\pgfplotstableread[row sep=\\,col sep=&]{
testset  &  MDiCoh &  EAGrid\\
UI	& 84.46 & 57.65 \\ 
UR	& 74.44 & 57.10 \\ 
EUO & 87.84 & 66.93 \\ 
    }\uo
\pgfplotstableread[row sep=\\,col sep=&]{
testset & MDiCoh &  EAGrid\\
UO  & 95.04 & 65.51\\
UI	& 88.20 & 60.93 \\ 
UR	& 74.91 & 55.59 \\ 
EUO & 85.24	& 65.33 \\ 
    }\ui
\pgfplotstableread[row sep=\\,col sep=&]{
testset  &  MDiCoh &  EAGrid\\
UO  & 92.75 & 58.70  \\
UI	& 82.14 & 52.65	\\ 		
UR	& 83.02 & 68.49	\\
EUO & 85.27 & 57.95	\\ 
    }\ur
\pgfplotstableread[row sep=\\,col sep=&]{
testset  &  MDiCoh &  EAGrid\\
UO  & 94.4 & 67.41 \\
UI	& 81.33 & 57.53 \\ 
UR	& 73.02 & 54.95 \\ 
EUO & 88.55 & 67.18 \\ 
    }\euo
\begin{figure*}[!t]
\resizebox{\textwidth}{!}{%
\begin{tabular}{@{}cccc@{}}
\begin{tikzpicture}
            \begin{axis}[
                    ybar,
                    bar width=5mm,
                    height=12cm,
                    legend style={at={(0.5,1.0)}, anchor=north,legend columns=-1,
                    font=\LARGE},
                    symbolic x coords={UO,UI,UR,EUO},
                    xtick=data,
                    nodes near coords,
                    every node near coord/.append style={rotate=90, anchor=west, font=\huge},
                    ymin=50,ymax=100,
                    ylabel={Acc (\%)},
                    ticklabel style= {font=\huge},
                    ylabel style={font=\huge}
                ]
                
        \addplot[fill=white] table[x=testset,y=MDiCoh]{\uo};
        \addplot[fill=black] table[x=testset,y=EAGrid]{\uo};
        \legend{MDiCoh, EAGrid};
        \end{axis};
        \end{tikzpicture}
         &
        \begin{tikzpicture}
            \begin{axis}[
                    ybar,
                    bar width=5mm,
                    height=12cm,
                    legend style={at={(0.5,1.0)}, anchor=north,legend columns=-1, font=\LARGE},
                    symbolic x coords={UO,UI,UR,EUO},
                    xtick=data,
                    nodes near coords,
                    every node near coord/.append style={rotate=90, anchor=west, font=\huge},
                    ymin=50,ymax=100,
                    ylabel={Acc (\%)},
                    ticklabel style= {font=\huge},
                    ylabel style={font=\huge}
                ]
        \addplot[fill=white] table[x=testset,y=MDiCoh]{\ui};
        \addplot[fill=black] table[x=testset,y=EAGrid]{\ui};
        \legend{MDiCoh, EAGrid};
        \end{axis}
        \end{tikzpicture}
         &
        \begin{tikzpicture}
            \begin{axis}[
                    ybar,
                    bar width=5mm,
                    height=12cm,
                    legend style={at={(0.5,1.0)}, anchor=north,legend columns=-1, font=\LARGE},
                    symbolic x coords={UO,UI,UR,EUO},
                    xtick=data,
                    nodes near coords,
                    every node near coord/.append style={rotate=90, anchor=west, font=\huge},
                    ymin=50,ymax=100,
                    ylabel={Acc (\%)},
                    ticklabel style= {font=\huge},
                    ylabel style={font=\huge}
                ]
        \addplot[fill=white] table[x=testset,y=MDiCoh]{\ur};
        \addplot[fill=black] table[x=testset,y=EAGrid]{\ur};
        \legend{MDiCoh, EAGrid};
        \end{axis}
        \end{tikzpicture}
         
         &
        \begin{tikzpicture}
            \begin{axis}[
                    ybar,
                    bar width=5mm,
                    height=12cm,
                    legend style={at={(0.5,1.0)}, anchor=north,legend columns=-1, font=\LARGE},
                    symbolic x coords={UO,UI,UR,EUO},
                    xtick=data,
                    nodes near coords,
                    every node near coord/.append style={rotate=90, anchor=west, font=\huge},
                    ymin=50,ymax=100,
                    ylabel={Acc (\%)},
                    ticklabel style= {font=\huge},
                    ylabel style={font=\huge}
                ]
        \addplot[fill=white] table[x=testset,y=MDiCoh]{\euo};
        \addplot[fill=black] table[x=testset,y=EAGrid]{\euo};
        \legend{MDiCoh, EAGrid};
        \end{axis}
        \end{tikzpicture}
         
         \\
         \huge{(a) \uoshort}
         &
         \huge{(b) \uishort}
         &
         \huge{(c) \urshort}
         &
         \huge{(d) \huoshort}
 \end{tabular}
 }
    \caption{Comparing \eagrid\ (black bars) and \mdicoh\ (white bars) in cross problem-domain. 
    The labels of figures are the perturbations of the training sets
    and the labels on x-axes are the perturbations of the test sets.}
    \label{fig:ext_results}
\end{figure*}
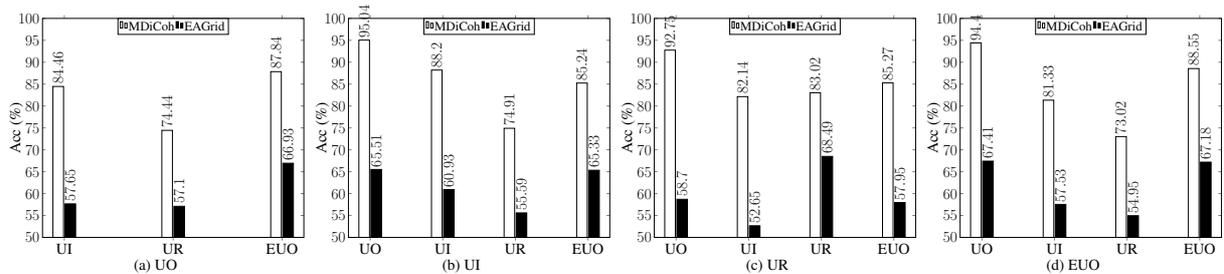

An explanation for why \mdicoh\ outperforms \aseq\ and \eagrid\ on \dd\ but not on \sw\ might be that  the \aseq\ and \eagrid\ models explicitly use \emph{gold} DA labels during evaluation but \mdicoh\ does not; and the DA \annotation s in \sw\ are about $10$ times higher fine-grained than those in \dd\ (see Table~\ref{tab:datasets}).
This interpretation becomes more concrete by observing a considerable reduction in the performance of \aseq\ and \eagrid\ when they are evaluated on \dd\ compared with when they are evaluated on \sw.
In contrast, our \mdicoh, which uses DAs only during training to obtain better utterance vectors, performs almost evenly on both corpora. 
%
Since our model does not need DA labels during evaluations, it is more suitable than the examined models for evaluating dialogue coherence in real scenarios.    

Finally, to shed some light on which parts of a dialogue receive higher attentions by our M-DiCoh model, we analyze the attention weights it assigns to utterance words.   
Table~\ref{tab:attention} illustrates the attention weights for an example dialogue from the training set of the \uoshort\ problem-domain on \dd, where words with higher attention weights are darker than the those with lower attention weights. 
\begin{table}[!b]
    \small
    \centering
    \begin{tabular}{@{}lp{5cm}l@{}}
        \toprule
         & Utterance & DA labels \\
        \midrule
        $utt_1$ &
        \textit{
         \textcolor{black!100}{hello} \textcolor{black!90}{,} \textcolor{black!100}{where} \textcolor{black!80}{can} \textcolor{black!70}{i}  \textcolor{black!50}{buy} \textcolor{black!40}{an} \textcolor{black!100}{inexpensive} \textcolor{black!90}{cashmere} \textcolor{black!50}{sweater} \textcolor{black!50}{?} 
         }
         & Question\\
         $utt_2$ &
        \textit{
         \textcolor{black!80}{maybe} \textcolor{black!60}{you} \textcolor{black!50}{should} \textcolor{black!50}{look} \textcolor{black!50}{around} \textcolor{black!50}{for} \textcolor{black!50}{an} \textcolor{black!100}{outlet} \textcolor{black!50}{.}
         }
         & Directive \\
         $utt_3$ &
         \textit{
         \textcolor{black!50}{that} \textcolor{black!50}{is} \textcolor{black!50}{a} \textcolor{black!50}{wonderful} \textcolor{black!80}{idea} \textcolor{black!50}{.} 
         }
         & 
         Commisive \\
         $utt_4$ &
         \textit{
         \textcolor{black!50}{outlets} \textcolor{black!50}{have} \textcolor{black!50}{more} \textcolor{black!50}{reasonable} \textcolor{black!90}{prices} \textcolor{black!50}{.} 
         }
         & 
         Inform\\
         $utt_5$ &
         \textit{
          \textcolor{black!50}{thank} \textcolor{black!50}{you} \textcolor{black!50}{for} \textcolor{black!50}{your} \textcolor{black!70}{help} \textcolor{black!50}{.}
          }
         & Inform\\
         $utt_6$ &
         \textit{
         \textcolor{black!50}{no} \textcolor{black!90}{problem} \textcolor{black!50}{.} \textcolor{black!90}{good} \textcolor{black!90}{luck} \textcolor{black!90}{.} 
         }
         & Inform\\
        \bottomrule
    \end{tabular}
    \caption{An illustration of attention weights assigned to words in a dialogue from \dd. Different gray shades show different attention weights.}
    \label{tab:attention}
\end{table}
We observe that using dialog act prediction as an auxiliary task helps our coherence model to assign high attention weights to the salient words in dialogue utterances.  
The wh-question, adjectives, and the verb in  questions have higher attention weights; while in other utterances, nouns, e.g.\ \emph{outlet}, \emph{inexpensive}, \emph{prices}, are more salient.  
So, our multi-task learning approach yields richer representations of dialog utterances for coherence assessment.   
\subsection{Cross Problem-domain Evaluation}
In a more challenging evaluation setup, we use the model trained on the training set of one \experiment\ to evaluate it on the test sets of the other \experiment s. 
Therefore, the perturbation methods used for constructing the training sets differ from those used for creating the test sets. 
%
%
We compare \eagrid\, as the state-of-the-art coherence model, and \mdicoh\, as our complete model, for cross problem-domain evaluations on \dd. 
%
\paragraph{Results}
Figure~\ref{fig:ext_results} shows the results on the test sets of the \experiment s, where the models are trained on the training set created by the (a) \uoshort, (b) \uishort, (c) \urshort, and (d) \huoshort\  perturbations. 
For all perturbations used to construct the training sets, we observe that \mdicoh\ outperforms \eagrid\ for all test perturbations.  
Interestingly, among all examined perturbations, both \mdicoh\ and \eagrid\ achieve the highest accuracy on \uoshort. 
We speculate that this perturbation is \mbox{easy-to-solve} as it re-arranges all utterances in a dialogue. 
\newcite{cervone18} also show that \urshort\ is easier to solve than \uishort . 

We note a low-discrepancy in the accuracy of the \mdicoh\ model on the test set of \uoshort\ when the model is trained on the training sets of the different examined \mbox{problem-domains}.  
The biggest drop in accuracy ($3.2$ percentage point) on the \uoshort\ \mbox{problem-domain} is for when the model is trained on the training set of the \urshort\ problem-domain. 
In contrast, we observe a high-discrepancy in the accuracy of the \eagrid\ model for the \uoshort\ problem-domain when the model is trained on the training sets of different problem-domains.   
The accuracy of \eagrid\ on the test set of \uoshort\ drops from $71.72\%$ (when trained for \uoshort ) to $58.7\%$ (when trained for \urshort ). 
This is about $13$ percentage points drop in accuracy. 
These results confirm that our \mdicoh\ model is more robust than the \eagrid\ model against different types of perturbation.  

\subsection{DAP Model Evaluation} 
Since using DAP as an auxiliary task improves the performance of our coherence model; in this experiment, we investigate the impact of MTL on the performance of the DAP model.  
We train our DAP model without any coherence supervision signal, S-DAP, with $\mathcal{L}=\frac{\mathcal{L}^{dial_i}_{da} + \mathcal{L}^{dial_j}_{da}}{2}$ in Equation~\ref{eq:total_loss}, and compare it with the model that is trained with our MTL, M-DAP. 

\paragraph{Results} Table~\ref{tab:da_dd} shows the F1 metric\footnote{We use F1 because there are more than two DA labels.} of these models for our \experiment s on the \dd\ dataset. 
This dataset is larger than \sw , and the frequency of dialogue act labels in this dataset is more balanced than those in \sw . 
\begin{table}[!t]
    \small
    \centering
    \resizebox{0.48\textwidth}{!}{
    \begin{tabular}{@{}lcccc@{}}
        \toprule
                     & \uoshort & \uishort & \urshort & \huoshort \\
        \midrule
        SVM-BoW    & $76.11$    &  $75.52$ & $74.49$ & $75.73$ \\ 
        \midrule
        S-DAP       & $78.10_{\pm .20}$ &  $79.15_{\pm .34}$ & $77.99_{\pm .35}$&  $78.81_{\pm .31}$\\ 
        M-DAP       & $77.32_{\pm .36}$ & $78.49_{\pm .33}$ & $77.52 _{\pm .27}$ & $78.51_{\pm .23}$\\
        \bottomrule
    \end{tabular}
    }
    \caption{The  F1 metric of the DAP model for the test sets of the \experiment s on \dd.  
    S-DAP is the model trained without any coherence supervision, and M-DAP is the model trained with MTL.} 
    \label{tab:da_dd}
\end{table}
We use an SVM classifier supplied with \mbox{Bag-of-Word} representations of utterances as a baseline to put our results in context. 

Both S-DAP and M-DAP models outperform the SVM-BoW model for all problem-domains, indicating that the employed DAP model is suitable for solving this task. 
However, we observe that the \mbox{M-DAP} model works on par with the \mbox{S-DAP} model. 
This observation shows that the information encoded by the coherence model is not useful for solving the dialogue act prediction task.   
The coherence model captures semantic relations in a dialogue by encoding information about the content of utterances. 
Dialogue acts, which indicate speakers' intentions of stating utterances in a dialogue, are independent of the content of utterances, therefore information learned by the coherence model does not help the DAP model. 

However, as the other experiments in this paper demonstrate, DAs can help to obtain more informative utterance representations to model dialogue coherence. 
Our multi-task learning approach relieves the need for explicit DA labels for coherence assessments,  which is the main goal of this paper. 

\section{Conclusions}
We propose a novel dialogue coherence model whose utterance encoder layers are shared with a dialogue act prediction model.
Unlike previous approaches that utilize these two models in a pipeline, we use them in a multi-task learning scenario where dialogue act prediction is an auxiliary task.  
Our coherence method outperforms its counterparts for discriminating dialogues from their various perturbations on \dd, and (mostly) performs on par with them on \sw. 
Our model (a) benefits from dialogue act prediction task during training to obtain informative utterance vectors, and (b) alleviates the need for gold dialogue act \annotation s during evaluations. 
These properties holistically make our model suitable for comparing different dialogue agents in terms of coherence and naturalness.    
For future work, we would like to deeply study the impacts of our perturbations on the coherence of the examined dialogues. 
We will also investigate to what extent the rankings of dialogues obtained by our model correlate with human-provided rankings. 

\section*{Acknowledgments}
This work was supported by the German Research Foundation through the German-Israeli Project Cooperation (DIP, grant DA 1600/1-1 and grant GU 798/17-1). 
We thank Kevin Stowe and 
Leonardo Filipe Rodrigues Ribeiro for their valuable feedback on earlier drafts of this paper. 
We also thank anonymous reviewers for their useful suggestions for improving the quality of the paper. 
\bibliography{lit,my_lit}

\begin{thebibliography}{34}
\expandafter\ifx\csname natexlab\endcsname\relax\def\natexlab#1{#1}\fi

\bibitem[{Barzilay and Lapata(2005)}]{barzilay05a}
Regina Barzilay and Mirella Lapata. 2005.
\newblock Modeling local coherence: An entity-based approach.
\newblock In \emph{Proceedings of the 43rd Annual Meeting of the Association
  for Computational Linguistics, {\em Ann Arbor, Mich., 25--30 June 2005}},
  pages 141--148.

\bibitem[{Barzilay and Lapata(2008)}]{barzilay08}
Regina Barzilay and Mirella Lapata. 2008.
\newblock Modeling local coherence: An entity-based approach.
\newblock \emph{Computational Linguistics}, 34(1):1--34.

\bibitem[{Burstein et~al.(2010)Burstein, Tetreault, and Andreyev}]{burstein10}
Jill Burstein, Joel Tetreault, and Slava Andreyev. 2010.
\newblock Using entity-based features to model coherence in student essays.
\newblock In \emph{Proceedings of Human Language Technologies 2010: The
  Conference of the North American Chapter of the Association for Computational
  Linguistics, {\em Los Angeles, Cal., 2--4 June 2010}}, pages 681--684.

\bibitem[{Byron and Stent(1998)}]{byron98}
Donna~K. Byron and Amanda Stent. 1998.
\newblock A preliminary model of centering in dialog.
\newblock In \emph{Proceedings of the 17th International Conference on
  Computational Linguistics and 36th Annual Meeting of the Association for
  Computational Linguistics, {\em Montr{\'e}al, Qu{\'e}bec, Canada, 10--14
  August 1998}}, pages 1475--1477.

\bibitem[{Cervone et~al.(2018)Cervone, Stepanov, and Riccardi}]{cervone18}
Alessandra Cervone, Evgeny Stepanov, and Giuseppe Riccardi. 2018.
\newblock \href {https://doi.org/10.21437/Interspeech.2018-2446} {Coherence
  models for dialogue}.
\newblock In \emph{Proceedings of the 19th Annual Conference of the
  International Speech Communication Association, {\em Hyderabad, 2--6
  September 2018}}, pages 1011--1015.

\bibitem[{Dinan et~al.(2019)Dinan, Logacheva, Malykh, Miller, Shuster, Urbanek,
  Kiela, Szlam, Serban, Lowe, Prabhumoye, Black, Rudnicky, Williams, Pineau,
  Burtsev, and Weston}]{dian18}
Emily Dinan, Varvara Logacheva, Valentin Malykh, Alexander~H. Miller, Kurt
  Shuster, Jack Urbanek, Douwe Kiela, Arthur Szlam, Iulian Serban, Ryan Lowe,
  Shrimai Prabhumoye, Alan~W. Black, Alexander~I. Rudnicky, Jason Williams,
  Joelle Pineau, Mikhail Burtsev, and Jason Weston. 2019.
\newblock \href {http://arxiv.org/abs/1902.00098} {The second conversational
  intelligence challenge (convai2)}.
\newblock \emph{CoRR}, abs/1902.00098.

\bibitem[{Dziri et~al.(2019)Dziri, Kamalloo, Mathewson, and Zaiane}]{dziri19}
Nouha Dziri, Ehsan Kamalloo, Kory Mathewson, and Osmar Zaiane. 2019.
\newblock Evaluating coherence in dialogue systems using entailment.
\newblock In \emph{Proceedings of the 2019 Conference of the North American
  Chapter of the Association for Computational Linguistics: Human Language
  Technologies, {\em Minneapolis, Minnesota., 2--7 June 2019}}, pages
  3806--3812.

\bibitem[{Farag and Yannakoudakis(2019)}]{farag19}
Youmna Farag and Helen Yannakoudakis. 2019.
\newblock Multi-task learning for coherence modeling.
\newblock In \emph{Proceedings of the 57th Annual Meeting of the Association
  for Computational Linguistics (Volume 1: Long Papers), {\em Florence, Italy,
  28 July -- 2 August, 2019}}, pages 629--639.

\bibitem[{Gandhe and Traum(2008)}]{gandhe08}
Sudeep Gandhe and David Traum. 2008.
\newblock Evaluation understudy for dialogue coherence models.
\newblock In \emph{Proceedings of the 9th SIGdial Workshop on Discourse and
  Dialogue, {\em Columbus, Ohio, 19--20 June 2008}}, pages 172--181.

\bibitem[{Gandhe and Traum(2016)}]{gandhe16}
Sudeep Gandhe and David Traum. 2016.
\newblock A semi-automated evaluation metric for dialogue model coherence.
\newblock In \emph{7th International Workshop on Spoken Dialogue Systems,
  \em{Saariselk{"a}, Finland, 13--16 January 2016}}, pages 141--150.

\bibitem[{Gao et~al.(2019)Gao, Meyer, Mesgar, and Gurevych}]{gaoyang19}
Yang Gao, Christian~M. Meyer, Mohsen Mesgar, and Iryna Gurevych. 2019.
\newblock \href {https://doi.org/10.24963/ijcai.2019/326} {Reward learning for
  efficient reinforcement learning in extractive document summarisation}.
\newblock In \emph{Proceedings of the 28th International Joint Conference on
  Artificial Intelligence, {\em Macao, China, 10--16 August 2019}}, pages
  2350--2356.

\bibitem[{Ghazvininejad et~al.(2018)Ghazvininejad, Brockett, Chang, Dolan, Gao,
  Yih, and Galley}]{ghazvininejad18}
Marjan Ghazvininejad, Chris Brockett, Ming~Wei Chang, Bill Dolan, Jianfeng Gao,
  Wen~Tau Yih, and Michel Galley. 2018.
\newblock A knowledge-grounded neural conversation model.
\newblock In \emph{Proceedings of the 32ed Conference on the Advancement of
  Artificial Intelligence, {\em New Orleans, Louisiana, 2--7 February 2018}},
  pages 5110--5117.

\bibitem[{Grosz and Sidner(1986)}]{grosz86}
Barbara~J. Grosz and Candace~L. Sidner. 1986.
\newblock Attention, intentions, and the structure of discourse.
\newblock \emph{Computational Linguistics}, 12(3):175--204.

\bibitem[{Guinaudeau and Strube(2013)}]{guinaudeau13}
Camille Guinaudeau and Michael Strube. 2013.
\newblock \href {http://www.aclweb.org/anthology/P13-1010.pdf} {Graph-based
  local coherence modeling}.
\newblock In \emph{Proceedings of the 51st Annual Meeting of the Association
  for Computational Linguistics (Volume 1: Long Papers), {\em Sofia, Bulgaria,
  4--9 August 2013}}, pages 93--103.

\bibitem[{Halliday and Hasan(1976)}]{halliday76}
M.~A.~K. Halliday and Ruqaiya Hasan. 1976.
\newblock \emph{Cohesion in English}.
\newblock London, U.K.: Longman.

\bibitem[{Jurafsky and Shriberg(1997)}]{jurafsky97}
Daniel Jurafsky and Elizabeth Shriberg. 1997.
\newblock Switchboard {SWBD-DAMSL} shallow-discourse-function annotation coders
  manual, draft 13.
\newblock Technical Report 97-02, University of Colorado at Boulder.

\bibitem[{Kendall et~al.(2018)Kendall, Gal, and Cipolla}]{kendall18}
Alex Kendall, Yarin Gal, and Roberto Cipolla. 2018.
\newblock Multi-task learning using uncertainty to weigh losses for scene
  geometry and semantics.
\newblock In \emph{Proceedings of the International Conference on Computer
  Vision and Pattern Recognition, {\em Salt Lake City, UT, 18--22 June 2018}},
  pages 7482--7491.

\bibitem[{Li et~al.(2017)Li, Su, Shen, Li, Cao, and Niu}]{liyanran17}
Yanran Li, Hui Su, Xiaoyu Shen, Wenjie Li, Ziqiang Cao, and Shuzi Niu. 2017.
\newblock Dailydialog: A manually labelled multi-turn dialogue dataset.
\newblock In \emph{Proceedings of the Eighth International Joint Conference on
  Natural Language Processing (Volume 1: Long Papers), {\em Taipei, Taiwan, 27
  November -- 1 December, 2017}}, pages 986–--995.

\bibitem[{Li et~al.(2019)Li, Kiseleva, and de~Rijke}]{zimingli19}
Ziming Li, Julia Kiseleva, and Maarten de~Rijke. 2019.
\newblock Dialogue generation: {F}rom imitation learning to inverse
  reinforcement learning.
\newblock In \emph{Proceedings of the 33rd Conference on the Advancement of
  Artificial Intelligence, {\em Honolulu, Hawaii, 21 January --1 February
  2019}}, pages 6722--6729.

\bibitem[{Liu et~al.(2016)Liu, Lowe, Serban, Noseworthy, Charlin, and
  Pineau}]{liuciawei16}
Chia-Wei Liu, Ryan Lowe, Iulian Serban, Mike Noseworthy, Laurent Charlin, and
  Joelle Pineau. 2016.
\newblock How {NOT} to evaluate your dialogue system: An empirical study of
  unsupervised evaluation metrics for dialogue response generation.
\newblock In \emph{Proceedings of the 2016 Conference on Empirical Methods in
  Natural Language Processing, {\em Austin, Texas, 1--5 November 2016}}, pages
  2122--2132.

\bibitem[{Mesgar and Strube(2014)}]{mesgar14}
Mohsen Mesgar and Michael Strube. 2014.
\newblock \href {http://www.aclweb.org/anthology/W14-3701.pdf} {Normalized
  entity graph for computing local coherence}.
\newblock In \emph{Proceedings of TextGraphs-9: Graph-based Methods for Natural
  Language Processing, Workshop at EMNLP 2014, {\em Doha, Qatar, 29 October
  2014}}, pages 1--5.

\bibitem[{Mesgar and Strube(2018)}]{mesgar18}
Mohsen Mesgar and Michael Strube. 2018.
\newblock A neural local coherence model for text quality assessment.
\newblock In \emph{Proceedings of the 2018 Conference on Empirical Methods in
  Natural Language Processing, {\em Brussels, Belgium, 31 October -- 4 November
  2018}}, pages 4328--4339.

\bibitem[{Pennington et~al.(2014)Pennington, Socher, and
  Manning}]{pennington14}
Jeffrey Pennington, Richard Socher, and Christopher Manning. 2014.
\newblock \href {http://www.aclweb.org/anthology/D14-1162} {Glove: Global
  vectors for word representation}.
\newblock In \emph{Proceedings of the 2014 Conference on Empirical Methods in
  Natural Language Processing, {\em Doha, Qatar, 25--29 October 2014}}, pages
  1532--1543.

\bibitem[{Perrault and Allen(1978)}]{perrault78}
C.~Raymond Perrault and James~F. Allen. 1978.
\newblock \href {https://www.aclweb.org/anthology/T78-1017} {Speech acts as a
  basis for understanding dialogue coherence}.
\newblock In \emph{Theoretical Issues in Natural Language Processing-2}.

\bibitem[{Purandare and Litman(2008)}]{purandare08}
Amruta Purandare and Diane~J. Litman. 2008.
\newblock \href {http://www.aaai.org/Library/FLAIRS/2008/flairs08-050.php}
  {Analyzing dialog coherence using transition patterns in lexical and semantic
  features}.
\newblock In \emph{Proceedings of the 21st International Florida Artificial
  Intelligence Research Society Conference, \em{Coconut Grove, Florida, 15--17
  May 2008}}, pages 195--200.

\bibitem[{Raheja and Tetreault(2019)}]{raheja19}
Vipul Raheja and Joel Tetreault. 2019.
\newblock Dialogue act classification with context-aware self-attention.
\newblock In \emph{Proceedings of the 2019 Conference of the North American
  Chapter of the Association for Computational Linguistics: Human Language
  Technologies, {\em Minneapolis, Minnesota., 2--7 June 2019}}, pages
  3727--3733.

\bibitem[{Reimers and Gurevych(2018)}]{reimers18}
Nils Reimers and Iryna Gurevych. 2018.
\newblock \href {http://arxiv.org/abs/1803.09578} {Why comparing single
  performance scores does not allow to draw conclusions about machine learning
  approaches}.
\newblock \emph{CoRR}, abs/1803.09578.

\bibitem[{Salton(1971)}]{salton71}
Gerard Salton. 1971.
\newblock \emph{The {S}{M}{A}{R}{T} Retrieval System -- Experiments in
  Automatic Document Processing}.
\newblock Englewood Cliffs, N.J.: Prentice Hall.

\bibitem[{Searle(1969)}]{searle69}
John Searle. 1969.
\newblock \emph{Speech Acts}.
\newblock Cambridge University Press, Cambridge, U.K.

\bibitem[{Serban et~al.(2016)Serban, Sordoni, Bengio, Courville, and
  Pineau}]{serban16}
Iulian~V Serban, Alessandro Sordoni, Yoshua Bengio, Aaron Courville, and Joelle
  Pineau. 2016.
\newblock Building end-to-end dialogue systems using generative hierarchical
  neural network models.
\newblock In \emph{Proceedings of the 30th Conference on the Advancement of
  Artificial Intelligence, {\em Phoenix, Arizona, 12--17 February 2016}}, pages
  3776--3783.

\bibitem[{Tien~Nguyen and Joty(2017)}]{nguyen17}
Dat Tien~Nguyen and Shafiq Joty. 2017.
\newblock \href {http://www.aclweb.org/anthology/P17-1121} {A neural local
  coherence model}.
\newblock In \emph{Proceedings of the 55th Annual Meeting of the Association
  for Computational Linguistics (Volume 1: Long Papers), {\em Vancouver,
  Canada, 30 July -- 4 August 2017}}, pages 1320--1330.

\bibitem[{Vakulenko et~al.(2018)Vakulenko, de~Rijke, Cochez, Savenkov, and
  Polleres}]{vakulenko18}
Svitlana Vakulenko, Maarten de~Rijke, Michael Cochez, Vadim Savenkov, and Axel
  Polleres. 2018.
\newblock Measuring semantic coherence of a conversation.
\newblock In \emph{Proceedings of the 17th International Semantic Web
  Conference, {\em Monterey, Ca., 8-12 October 2018}}, pages 634--651.

\bibitem[{Xu et~al.(2018)Xu, Du{\v{s}}ek, Konstas, and Rieser}]{xuxinnuo18}
Xinnuo Xu, Ond{\v{r}}ej Du{\v{s}}ek, Ioannis Konstas, and Verena Rieser. 2018.
\newblock Better conversations by modeling, filtering, and optimizing for
  coherence and diversity.
\newblock In \emph{Proceedings of the 2018 Conference on Empirical Methods in
  Natural Language Processing, {\em Brussels, Belgium, 31 October -- 4 November
  2018}}, pages 3981--3991.

\bibitem[{Zhang et~al.(2018)Zhang, Lan, Guo, Xu, and Cheng}]{zhanghainan18}
Hainan Zhang, Yanyan Lan, Jiafeng Guo, Jun Xu, and Xueqi Cheng. 2018.
\newblock Reinforcing coherence for sequence to sequence model in dialogue
  generation.
\newblock In \emph{Proceedings of the 27th International Joint Conference on
  Artificial Intelligence, {\em Stockholm, Sweden, 13--19 July 2018}}, pages
  4567--4573.

\end{thebibliography}
\bibliographystyle{acl_natbib}

\clearpage
\appendix
\appendix
\section{More Details on EAGrid}
EAGrid is a recent model for dialogue coherence with which we compare our models. 
It mainly extends the entity grid representation for monologue texts.  
Entity grid is a matrix whose rows represent dialogue utterances and columns encode entities mentioned in dialogue.
Each entry in an entity grid is filled by the grammatical role (i.e.\ subject (``\texttt{S}''), object (``\texttt{O}''), neither of them (``\texttt{X}'')) of its corresponding entity in its corresponding utterance if the entity is mentioned in the utterance, otherwise it is filled by ``\texttt{\--}''. 
EAGrid appends a column for encoding dialogue acts to the entity grid such that the entries associated with this column are filled by the dialog act labels of corresponding utterances. 
Figure~\ref{fig:egrid} shows the EAGrid representation of the example dialogue presented in the top-part of Table 1 in this paper.  
The grid is generated using EAGrid's code released by its authors. 
\begin{figure}[!hb]
    \small
    \centering
    \resizebox{0.5\textwidth}{!}{
    \begin{tabular}{cccccc}
    \toprule
    & \multicolumn{4}{c}{Entities} & DA labels\\
    &  CHARLES & CAPTAIN & UNCLE & THIS  \\
    $utt_1$ & \texttt{X} & \texttt{\_} & \texttt{X} & \texttt{S} & inform \\
    $utt_2$ & \texttt{\_} & \texttt{\_} & \texttt{\_} & \texttt{\_} & question \\
    $utt_3$ & \texttt{\_} & \texttt{X} & \texttt{\_} & \texttt{\_} & inform \\
    $utt_4$ & \texttt{\_} & \texttt{\_} & \texttt{\_} & \texttt{\_} & inform \\
    $utt_5$ & \texttt{\_} & \texttt{\_} & \texttt{\_} & \texttt{\_} & inform \\
    \bottomrule
    \end{tabular}
    }
    \caption{ The EAGrid representations of Dialogue presented in Table .}
    \label{fig:egrid}
\end{figure}
The probabilities of entities' grammatical role and dialogue act label transitions of length $n$ across utterance are used as coherence features\footnote{Following the EAGrid model, we set $n=2$.}. 
These features are supplied to a \textit{SVM} to rank dialogues concerning their coherence. 
\section{LSTM}
As the LSTM layer used in our model is well-known, we give the details of its definition here: 
%
%
\begin{equation}
    \begin{split}
        i_t & = \sigma(W_{ii}e_t+b_{ii}+ W_{hi}h_{(t-1)}+b_{hi}),\\
        f_t & = \sigma(W_{if}e_t+b_{if}+ W_{hf}h_{(t-1)}+b_{hf}),\\
        g_t & = \tanh{(W_{ig}e_t+b_{ig}+ W_{hg}h_{(t-1)}+b_{hg})},\\
        o_t & = \sigma(W_{io}e_t+b_{io}+ W_{ho}h_{(t-1)}+b_{h0}),\\
        c_t & = f_t*c_{(t-1)} + i_t *g_t, \\
        h_t &= o_t*\tanh{(c_t)},
    \end{split}
    \label{eq:lstm-details}
\end{equation}
where $h_t$, is the hidden and $c_t$ is the cell state at word $t$. 
The input, forget, cell, and output gates at word $t$ are shown by $i_t$, $f_t$, $g_t$, and $o_t$, respectively. 
$\sigma$ is the Sigmoid function, and $*$ is the Hadamard product. 
The hidden state is initialized with a zero vector for representing each utterance in dialogue. 
\section{Hyperparameters and Training}
To approximate the best values of the hyperparameters, we perform a grid search, in which one parameter is varied while all others are fixed. 
The search was carried out in the multi-task learning setup on the dataset for the \uoshort\ problem-domain on \dd . 
For each variation of hyperparamter values, we train the model on the training set of \uoshort\ and evaluate it on its validation set. 
The parameter values that result in the highest respective performance was chosen for evaluation on the test set. 
The values for the number of epochs and batch size were chosen to trade off the running time and memory consumption of the training. 
For the experiments on \dd\ we set the maximum number of epochs to 20 and the batch size to 128, while for \sw\ the maximum number of epochs is set to 10 and the batch size to 16. 
Note that hyperparameter tuning has not been performed for \sw . 
Thus for the experiments on \sw\ mostly the same hyperparameters as those used for the experiments on \dd\ are used, with the exception of the batch size and the number of epochs.
Table~\ref{tab:hype} shows the final values for hyperparameters of our models. 
The optimization is performed by Adam with its default parameter values except for the learning rate. 
We train the model on the shuffled batches of training data. 
The model is evaluated on the validation set at each epoch. 
The model with the best performance on the validation set is chosen for the evaluation on the test set. 
The training procedure is accelerated by the usage of a Tesla P100 GPU running with CUDA v.10.1, while the model is implemented in the pytorch7 framework version 1.1.0.
\begin{table}[!ht]
    \small
    \centering
    \begin{tabular}{lcc}
         \toprule
         parameter& \dd\ & \sw\  \\
         \midrule
          epochs & 20 & 10 \\
          batch size & 128 & 16 \\
          learning rate & 0.0005 &  0.0005\\
          number of LSTM layers & 1 & 1\\
          hidden layer of LSTM$_u$ & 128& 128 \\
          hidden layer of LSTM$_d$ & 256& 256 \\
          DA dropout rate  & 0.1& 0.1\\ 
         \bottomrule
    \end{tabular}
    \caption{The values of hyperparameters that result in the best performance on the validation set.}
    \label{tab:hype}
\end{table}

\end{document}